%% file: mtdeep.tex
\documentclass[runningheads]{llncs}
\usepackage[dvipsnames,table]{xcolor}
\usepackage{amsfonts}       
\usepackage{enumitem}
\usepackage{balance}
\usepackage{multirow,array}
\usepackage{tikz}
\usepackage{pgfplots}
\usetikzlibrary{pgfplots.dateplot}
\usepackage{pgfplotstable}
\usepackage{filecontents}
\usetikzlibrary{backgrounds}
\graphicspath{ {images/} }
\usepackage{wrapfig}
\usepackage{fontawesome}
\usepackage{wrapfig}
\usepackage{graphicx}

\usepackage{silence}
\usepackage{subcaption}
\begin{document}
	\title{MTDeep: Boosting the Security of Deep Neural Nets Against Adversarial Attacks with\\Moving Target Defense}
	%
	%
	\author{
		Sailik Sengupta\inst{1} \and
		Tathagata Chakraborti\inst{2} \and
		Subbarao Kambhampati\inst{1}\thanks{To be presented at the Conference on Decision and Game Theory for Security, 2019}
	}
	\titlerunning{MTDeep: Moving Target Defense for Deep Neural Networks}
	
	\authorrunning{Sengupta {\em et. al.}}
	%
	\institute{Arizona State University, Tempe, AZ, USA\\
		\email{\{sailiks,rao\}@asu.edu} \and
		IBM Research, Cambridge, MA, USA\\
		\email{tchakra2@ibm.com}
	}
	
	\maketitle              
	\begin{abstract}
		
		Present attack methods can make state-of-the-art classification systems based on deep neural networks mis-classify every adversarially modified test example.
		The design of general defense strategies against a wide range of such attacks still remains a challenging problem.
		In this paper, we draw inspiration from the fields of cybersecurity and multi-agent systems and propose to leverage the concept of {\em Moving Target Defense (MTD)} in designing a meta-defense for `boosting' the robustness of an ensemble of deep neural networks (DNNs) for visual classification tasks against such adversarial attacks.
		To classify an input image, a trained network is picked randomly from this set of networks by formulating the interaction between a Defender (who hosts the classification networks) and their (Legitimate and Malicious) users as a \emph{Bayesian Stackelberg Game (BSG)}.
		We empirically show that this approach, {\em MTDeep}, reduces misclassification on perturbed images in various datasets such as MNIST, FashionMNIST, and ImageNet while maintaining high classification accuracy on legitimate test images.
		We then demonstrate that our framework, being the first meta-defense technique, can be used {\em in conjunction} with any existing defense mechanism to provide more resilience against adversarial attacks that can be afforded by these defense mechanisms. Lastly, to quantify the increase in robustness of an ensemble-based classification system when we use MTDeep, we analyze the properties of a set of DNNs and introduce the concept of differential immunity that formalizes the notion of attack transferability.
	\end{abstract}

\input{data/imagenet.tex}\input{data/mnist.tex}\input{data/at_mnist.tex}

\input{data/fashion_mnist.tex}
	\section{Introduction}
	State-of-the-art systems for image classification based on Deep Neural Networks (DNNs) are used in many important tasks such as recognizing handwritten digits on cheques \cite{jayadevan2012automatic}, object classification for automated surveillance \cite{javed2006tracking} and autonomous vehicles \cite{de1997road}.  
	Adversarial attacks to make these classification systems misclassify inputs can lead to dire consequences.
	For example, in \cite{papernot2016limitations}, road signs saying `stop' are misclassified, which can make an autonomous vehicle behave dangerously. 
	Such attack mechanisms also exist for state-of-the-art vision systems that recognize faces, which may be used for authentication, target identification etc. as shown in \cite{sharif2016accessorize}.
	Moreover, one desires these adversarially generated images to be indistinguishable wrt the original image when evaluated from the perspective of a human classifier.
	Thus, if $\hat{D}(i)$ denotes the class of an image $i$ output by a Deep Neural Network $\hat{D}$, an adversarial perturbation $\epsilon$ when added to the image $i$ tries to ensure that $\hat{D}(i) \neq \hat{D}(i+\epsilon)$.  In addition, attackers try to minimize some norm of $\epsilon$, which ensures that the changed image $i+\epsilon$ and the original image $i$ are indistinguishable to humans.
	The effectiveness of an attack method is measured by the accuracy of a classifier on the perturbed images generated by it.
	
	Defenses against adversarial examples are designed to be effective against a certain class of attacks by either training the classifier with perturbed images generated by these attacks or making it hard for these attacks to modify some property of the neural network. 
	Unfortunately, this has evolved into a cat-and-mouse game and often, a state-of-the-art defense mechanism is proved to be inadequate against a new class of attacks almost as soon as it is published. Some recent works try to formulate the attack scenario as a min-max adversarial game where the defender tries to minimize the loss while the attacker tries to maximize it. They show the use of Projected Gradient Descent (PGD) for solving the inner (max) optimization can result in attacks that are extremely effective in crippling the classification system and, at the same time, capture the characteristics of many of state-of-the art attacks \cite{madry2017towards}. They claim that robust training methods that enforce classification to the same class when images are $\epsilon$ distance away from any image in the training set results in high classification accuracy against adversarial examples. Unfortunately, this has the side effect of bringing down the classification accuracy on non-perturbed examples.
	
	\input{figs/mtdeep_concept.tex}
	
	In this paper, we take a different view and try to design a meta-defense that can function both as (1) a first line of defense against new attacks and (2) a second line of defense used in conjunction with any existing defense mechanism to boost the security gain the latter can provide.
	To this end, we take a game theoretic perspective and investigate the use of Moving Target Defense (MTD) \cite{zhuang2014towards}, in which we randomly select a network from an ensemble of networks when classifying an input image (randomization at test time), for boosting the robustness against adversarial attacks (see Fig. \ref{fig:mtdnn}). Our contributions are--
	
	
	\begin{itemize}
		\item[--]
		MTDeep -- an MTD framework for an ensemble of DNNs, which can be used as a meta-level defense-in-depth mechanism, to bootstrap any existing defense mechanism and increase the robustness of the classification system to different classes of adversarial attacks.\\
		
		\item[--]
		A Bayesian Stackelberg Game formulation with two players-- MTDeep and the users. We then solve a multi-objective optimization problem to obtain the Stackelberg Equilibrium of this game, which gives us the optimal randomization strategy for the ensemble and also maximizes the classification accuracy on regular as well as adversarially modified inputs.\\
		
		\item[--]
		Empirical evaluation to show that MTDeep can be used as (1) a standalone defense mechanism to increase the accuracy on adversarial samples by $\approx 24\%$ for MNIST, $\approx 22\%$ for Fashion MNIST and $\approx 21\%$ for ImageNET data-sets against a variety of well-known attacks and (2) in conjunction with existing defense mechanisms like Ensemble Adversarial Training, MTDeep increases the robustness of a classification system (by $\approx 50\%$ for MNIST). We also show that black-box attacks (c.f. related work) on a distilled networked are ineffective (in comparison to white-box attacks) against the MTDeep system.\\
		
		\item[--]
		Analysis on an ensemble of DNNs for MNIST data that elucidates how much of a security gain MTDeep can provide. As a part of that analysis, we define the concept of {\em differential immunity}, which is (1) the first attempt at defining a robustness measure for an ensemble against attacks and (2) a quantitative metric to capture the notion of attack transferability.
	\end{itemize}
	
	Although prior research has shown that effectiveness of attacks can sometimes transfer across networks \cite{szegedy2013intriguing}, we show that there is still enough residual disagreement among networks that can be leveraged to design an add-on defense-in-depth mechanism by using MTD. 
	In fact, recent work has demonstrated that 
	it is possible to train models with limited adversarial attack transferability \cite{adam2018stochastic}, making our meta-level defense approach particularly attractive.
	
	Our work is also different from any previous work that uses ensembles to defend against attacks. In general, the concept of DNN-based ensembles simply tries to increase classification accuracy for legitimate users but provides no protection against adversarial examples \cite{ioffe2015batch}.
	In \cite{abbasi2017robustness}, researchers propose an ensemble-based method to detect adversarial samples for the MINST dataset (and this can be extended when the constituent networks are DNNs).
	Unfortunately, such voting based mechanisms for ensembles can be viewed as simply adding an extra pooling layer whose weights are equal to the importance given to the votes of the constituent networks.
	At this point, all attacks on a DNNs are trivially effective against such these voting-based ensembles.  Furthermore, researchers have also shown that an ensemble of vulnerable DNNs cannot result in a classifier robust to attacks \cite{he2017adversarial}.
	In contrast, MTDeep builds in an implicit mechanism based on {\em randomization at prediction time}, making it difficult for an adversary to fool the classification system.

	\section{Related Work}
	\label{sec:rw}
	
	\subsection{Attacks and Defenses for Deep Neural Networks}
	
	In this section, we briefly review existing work on crafting adversarial inputs against deep neural networks (at test-time) and defenses developed against them.
	
	\paragraph{Gradient-based perturbations:} 
	Recent literature has shown multiple ways of crafting adversarial samples for a DNN \cite{moosavi2016deepfool,papernot2016limitations,szegedy2013intriguing,madry2017towards}. 
	In these works, either (i) the input features whose partial derivatives on the DNN's loss functions are high are modified by a small amount to maximize the DNN's loss function and therefore make the classifier misclassify them, or (ii) the geometric space around a point is examined to find the closest class-separation boundary and generate perturbation vectors that push the modified image to the other side of this boundary.  
	Similar to a chosen ciphertext attack, these attacks assume that the test image which is to be modified is available beforehand. Furthermore, they assume availability of complete knowledge about the classification network.
	\paragraph{Black-box attacks:} 
	Black-box attacks against DNNs train a small substitute model by assuming that the network being attacked provides test labels for a list of images the adversary provides \cite{papernot2017practical}, similar to chosen plaintext attacks. Surprisingly, attacks on this substitute model generalize to the actual network \cite{szegedy2013intriguing}. Recent work on zeroth-order optimization has shown it is possible to create black-box attacks without the need for substitute models \cite{chen2017zoo}.
	
	\paragraph{Defenses.} Defense techniques against the two types of attacks described above commonly involve (1) generating adversarial perturbed training images using one (or all) of the attack methods described and then (2) using the generated images along with the correct labels to fine tune the parameters of the DNN during training. 
	This helps the DNN to correct its bias in some of the unexplored areas of the high dimensional space, reducing the effectiveness of the adversarial perturbations. Ensemble adversarial training \cite{tramer2017ensemble} and stability training \cite{zheng2016improving} are two improvements on this defense technique.\footnote{\small Ensemble Adversarial Training, uses an ensemble to generate adversarially perturbed examples for all the constituent networks and uses them to strengthen a single network \cite{tramer2017ensemble}. Unlike us, it does not use the ensemble at classification time.}
	Besides these, researchers have developed defense mechanisms like gradient masking \cite{papernot2016towards}, defensive distillation \cite{papernot2016distillation} and dimensionality reduction \& `anti-whitening' \cite{DBLP:journals/corr/BhagojiCM17}. Some of these defense mechanisms, similar to trends in cybersecurity, have been rendered ineffective due to discovery of stronger attacks (eg. \cite{carlini2017towards}). We do not consider these methods further since {\em our proposed framework can be used in conjunction with any of these to improve their security guarantees}.
	Lastly, our approach is further supported by previous research works that show introduction of randomized switching makes it harder for any attacker to reverse engineer a classification system with precision \cite{vorobeychik2014optimal}, which is how most white-box attacks are constructed.
	\vspace{-4pt}
	\paragraph{Universal perturbations:} 
	This DNN-specific perturbation when added to any input image, makes a DNN misclassify it \cite{moosavi2016universal}.
	This attack is based on the DeepFool attack \cite{moosavi2016deepfool} and although it is often time consuming to generate, only one ``universal'' perturbation per network needs be computed. Moreover, authors show that adversarial training is ineffective in increasing robustness against these attacks. Also, other state-of-the-art defense mechanisms (\cite{tramer2017ensemble,madry2017towards}) have not shown that they can mitigate this attack. Newer class of such DNN-specific attacks such as Adversarial Patches \cite{brown2017adversarial} and BadNets \cite{gu2017badnets} relax the constraint that the perturbation is imperceivable to a human. We show that randomly switching between networks of an ensemble to classify an input image, as MTDeep does, can prove to be an effective defense against these attacks because these attacks are network specific and often have low transferability.
	
	There has been some effort in trying to protect machine learning systems against attacks using randomization techniques at test time \cite{biggio2008adversarial}. Unfortunately, these are not general enough to be used for DNNs. Furthermore, these mechanisms try to prevent misclassification rate under attack and end-up affecting the classification accuracy on non-adversarial or legitimate test inputs.
	
	\subsection{Moving Target Defense}
	
	Moving Target Defense (MTD) is a paradigm used in software security that tries to reduce the success rate of an attack by pro-actively switching between multiple software configurations \cite{zhuang2014towards}, thus enhancing system security \cite{taguinod15:mtd-web-app}.
	Based on the principles of MTD, we design a general purpose security framework for Deep Neural Networks in this paper.
	
	Devising effective switching strategies for MTD systems requires reasoning about attacks in a multi-agent game theoretic fashion in order to provide formal guarantees about the security of such systems \cite{paruchuri2008playing}. In  \cite{sengupta2017game}, the authors provide a Bayesian Stackelberg Formulation of the problem and show that the optimal mixed strategy leads to defense strategies that outperform trivial randomization strategies.
	Thus, we model the interaction between an image classification system driven by a ensemble of DNNs (MTDeep) and its users, both legitimate and adversarial, as a Bayesian Stackelberg Game, providing provable guarantees on the expected performance on \textit{both} legitimate and adversarial inputs, a consideration absent in the works on defense mechanism design for classifiers in general and DNNs in particular.

	\section{MTDeep: Moving Target Defense for Deep Neural Networks}
	\label{sec:mtdnn}
	
	In a Moving Target Defense (MTD) system, the defender has multiple system configurations.  The attacker has a set of attacks that it can use to cripple the configurations in the defender's system.
	Given an input to the system, the defender selects, {\em at random}, one of the configurations to run the input and returns the output generated by that system. 
	Since the attacker does not know which system was specifically selected, its attacks are less effective than before (Figure \ref{fig:mtdnn}).
	As stated earlier, randomization in selecting a configuration for classification of each input is paramount.  Unfortunately, an MTD framework for classification systems, that leverages randomimzation, might end up reducing the accuracy of the overall system in classifying non-perturbed images, because the DNN that has the highest accuracy is not always used for classification.  Thus, in order to retain good classification accuracy and guarantee high security, we model the interaction between MTDeep and the users as a Bayesian Stackelberg Game. We show that the equilibrium results in the optimal selection strategy.
	
	
	
	\subsection{The Defender}
	
	The defender (MTDeep) provides a service for classification of images. The configuration space for MTDeep are the DNNs in the ensemble that are trained on the particular image classification task.
	Let $N$ represent the set of defender configurations.
	In the ensemble we design for our experiments on MNIST and Fashion-MNIST datasets, we have three networks $N=\{$CNN, MLP, HRNN$\}$. The networks, evident from their names, are based on three different network architectures-- Convolution Neural Networks (CNN) \cite{krizhevsky2012imagenet}, Multi-Layer Perceptrons (MLP) and Hierarchical Recurrent Neural Networks \cite{du2015hierarchical}-- all of which give reasonably high accuracy on the two data-sets (Figure \ref{table:MNIST_LA}). For experiments on the ImageNET dataset, we use six pre-trained networks that have won the image classifications over the last few years and have a reasonable high accuracy on the data-set (see Table \ref{table:A}).
	It is worth emphasizing that this ensemble of classifiers does not behave like well-known voting based ensembles and at classification time, uses only a single network's decision.
	Formally, a pure strategy for the defender corresponds to selecting a single constituent network in the ensemble for each test input and use it for classification. A mixed strategy is a probability distribution over the different pure strategies and on every test input, the defender rolls a die (that represents the mixed strategy) to determine the constituent network it will use for classification.
	
	\input{tables/gm_f_n_mnist.tex}
	\input{tables/gm_imagenet.tex}
	
	\subsection{The User}
	
	The second player in this game is the user of the classification system. We divide the user into two player types--Legitimate User ($\mathcal{L}$) and the Adversary ($\mathcal{A}$).  $\mathcal{L}$ tries to input non-perturbed images to MTDeep system for classification, which is its only action.  
	The second type is the adversary $\mathcal{A}$ who essentially tries to perturb input images such that the classification system misclassifies these inputs.
	In our threat model, we consider a strong adversary who knows the different architectures we use in our MTDeep system. This means they can easily generate powerful white-box attacks for each of the networks in our system. We let $U$  denote this set of attacks the attacker can generate against our system. For MNIST and Fashion-MNIST, we consider three classes of white-box attacks-- the Fast Gradient Method (FGM), the DeepFool (DF) attack and the Projected Gradient Descent.\footnote{The reward for PGD attacks are whole numbers because, in order to compute attack-based perturbations in a reasonable amount of time, we evaluate its effectiveness for perturbing 100 samples (as opposed to 10000 in the case of FGM and DF).}
	(PGD) attacks (see Table \ref{table:MNIST_LA}) while for ImageNET, we restrict ourselves to universal perturbation attacks because the cost of constructing perturbed images for each test image is computationally intensive (see Table \ref{table:A}). 
	Each attack ($u \in U$) is generated to cripple a particular constituent network in the MTDeep ensemble (indicated using the sub-script) but used against each of the defender's constituent networks. They may or may not be equally effective for all the configurations. In fact, for most the white-box attacks such as FGM and PGD, we show that although theyhave some {\em transferability} across the different networks, no attack can completely cripple all the networks (Table  \ref{table:MNIST_LA} and \ref{table:A}). MTDeep, as we will later see, leverages this fact to boost the security against adversarial examples. 
	
	\subsection{Bayesian Game}
	
	MTDeep randomly picks a network $n$ ($\in N$) each time to classify an input image. If we use a naive switching strategy, such as uniform random selection, to pick a network whenever an input is provided, we will have equal chances of choosing networks that have (1) low classification accuracy or (2) high vulnerability to perturbed images, which will be sub-optimal. Also, the attacker might eventually infer the defender's switching strategy and exploit the highly vulnerable configurations more often. Thus, reasoning about the optimal strategy becomes necessary.
	Furthermore, as shown later, arbitrary switching strategies for an ensemble can be worse than using single networks in terms of classification accuracy. 
	Thus, we model the interaction between the ensemble and the users as a Bayesian Game, which helps us compute the optimal switching strategy.
	
	Existing works that design defense methods against adversarial attacks for DNNs model the problem as a zero-sum game where the attacker tries to maximize the defender's loss function by coming up with perturbed test examples that the network misclassifies, whereas the defender tries to reduce the loss on these adversarially perturbed examples. Fine tuning the classifier to have high accuracy on adversarially perturbed inputs often has the side effect of reducing the classification accuracy on non-perturbed inputs from the test set \cite{madry2017towards}.
	In this paper, we move away from the zero-sum game assumption and try to ensure that the defender minimizes the loss functions for both types of inputs images-- original test set images and the adversarially perturbed ones.  Thus, we want MTDeep to be effective for $\mathcal{L}$ (proportional to minimizing the loss on the original test set) and, at the same time, increase the accuracy of classification for the perturbed images (proportional to minimizing the loss against adversarial inputs at test-time), making this a multi-objective optimization problem.
	
	A natural question that arises is how much importance should we associate with the two different objectives. Unfortunately, this is highly application specific. For example, a banking system that uses handwritten digit recognition for identifying monetary amount on bank cheques should prioritize maximizing accuracy on adversarial examples over an occasional misclassification on the actual test set, whereas, an image captioning system that is trying to help a visually challenged person understand posts on social media hardly needs to care about adversarial examples. We capture this trade-off as the probability of the defender's belief about whether a particular input at test time is drawn from the legitimate test set or is an adversarially perturbed image, making this a Bayesian game. The utilities of each player in this game are as follows.
	
	\begin{itemize}
		\item[--]
		The Legitimate User ($\mathcal{L}$) and the defender both get a reward value that represents the accuracy of the DNN system.  Thus, for using a network $n$ in the ensemble $N$ with classification accuracy (say) $98\%$ for an input image both the defender and $\mathcal{L}$ get a reward of $98$ (see Tables \ref{table:MNIST_LA} \& \ref{table:A}).\\[-1ex]
		
		\item[--]
		The Adversary ($\mathcal{A}$) and the defender play a constant($=100$) sum game, where the former's reward value for an attack $u$ against the network $n$ is given by $e_{n,u}$, which is the {\em fooling rate} and the defender's reward is the accuracy of $n$ on perturbed inputs, which is $(100-e_{n,u})$ (Figure \ref{table:MNIST_LA}).
	\end{itemize}
	
	\subsection{MTDeep's Switching Strategy}
	
	Note that the defender $D$ has to play first, i.e. deploy a classification system that either a legitimate user $\mathcal{L}$ can use or an adversary $\mathcal{A}$ can attack. This imparts a leader-follower paradigm to the formulated Bayesian Game. The defender leads by playing first and then the attacker follows by choosing an attack action having inferred the leader's (mixed) strategy.  Satisfying the multi-objective criterion, mentioned above, is now equivalent to finding the Stackelberg Equilibrium of this game.
	This equilibrium can be found using the mixed integer quadratic program (MIQP) formulated in \cite{paruchuri2008playing}, which we now briefly describe. Let us denote the strategy vector for the defender as $\vec{x}$ and their reward as $R^{D}_{n,u}$ when the defender uses the network $n$ and user selects the action $u$. Similarly, the strategy vectors for the adversary and the legitimate user types are $\vec{q}^{\mathcal{A}}$ and $\vec{q}^{\mathcal{L}}$ and their rewards are $R^{\mathcal{A}}_{n,u}$ and $R^{\mathcal{L}}_{n,u}$ respectively.
	We seek to maximize the defender's reward while allowing the attacker to choose the most effective attack.
	Specifically, we solve the following optimization problem --
	
	\vspace{-10pt}
	\begin{eqnarray}
		\max_{x,q} \sum\limits_{n\in N}^{}(\alpha \cdot \sum\limits_{u \in U}^{} R_{n,u}^D~x_n q_u^{\mathcal{A}} &+& (1-\alpha) \cdot R_{n,u}^D~x_n q_u^{\mathcal{L}}) \nonumber \label{equation:o}\\
		\textnormal{s.t.}~~~ \sum\limits_{n\in N}^{} x_n &=& 1  \nonumber \\
		\sum\limits_{u\in U}^{} q_u^{\mathcal{Y}} &=& 1 ~~~~~\forall~ \mathcal{Y} \in \{\mathcal{A},\mathcal{L}\} \nonumber \\
		0 \leq &x_n& \leq 1 ~~~~~\forall~ n \in N \nonumber \\
		q^{\mathcal{Y}}_u &\in& \{0,1\} \nonumber ~~~~~\forall~ \mathcal{Y} \in \{\mathcal{A},\mathcal{L}\} \\
		0 \leq v^{\mathcal{Y}} - \sum\limits_{n\in N} R_{n,u}^\mathcal{Y} x_n  &\leq& (1-q_a^{\mathcal{Y}})M \nonumber \\ 
		\forall~u \in U^\mathcal{Y} && ~\forall~ \mathcal{Y} \in \{\mathcal{A},\mathcal{L}\} \nonumber
	\end{eqnarray}%
	
	\noindent where $\alpha$ is the probability of $\mathcal{A}$ attacking a MTDeep system and $M$ is a large positive number.
	The objective function maximizes the defender's expected reward over its own switching strategy $\vec{x}$ and the strategy vector played by the two user types ($\vec{q}^\mathcal{A},\vec{q}^\mathcal{L}$) weighted by their relative importance $\alpha$, which is the probability with which the defender expects the attacker type $\mathcal{A}$ attacks their system.
	
	Thus, this MIQP, implemented in Gurobi, takes as input (1) the reward values $R^D, R^\mathcal{A},$ and $R^\mathcal{L}$ obtained from accuracy metrics of the constituent networks, (2) $\alpha$, the probability of the player types and outputs the optimal strategy for both the defender ($\vec{x}$) and the users.
	The first four constraints ensure that the strategy vectors sum up to one since they represent probability of selecting actions.
	The fifth constraint represents the dual of the attacker's optimization problem which tries to maximize their expected reward $v^\mathcal{Y}$ over the defender's strategy. This constraint captures the fact that the attacker knows $\vec{x}$ and uses it to select its attack strategy $\vec{q}^\mathcal{A}$.
	Note that the second constraint forces the users $\mathcal{L}$ and $\mathcal{A}$ to select a pure strategy. As the authors in \cite{paruchuri2008playing} show, this constraint is not limiting for the attacker because for the attacker $\mathcal{A}$, there always exists a pure strategy in support of any mixed strategy it can play.
	For the attack the attacker selects, the right side of the fifth constraint becomes $0$ making $v^\mathcal{Y} = \sum_{n\in N} R_{n,u}^\mathcal{Y}$.
	Lastly, the defender's strategy, in the worst case, can be a pure strategy that directs MTDeep to use a single network for classification.
	
	\section{Experimental Results}
	\label{sec:er}
	
	We first compare the effectiveness of MTDeep for MNIST, Fashion-MNIST and the ImageNet datasets when it is used as a standalone defense mechanism.
	We then show that MTDeep when piggybacked onto an existing defense mechanism like Ensemble Adversarial Training, can result in boosting the accuracy against adversarial attacks. We then analyze the effect of black-box attacks designed using a distilled network that can capture a holistic view of an ensemble like MTDeep which leverages randomization at test time. We show that given the limiation on the number of samples in the of the MNIST dataset, blackbox attacks are less effective than white box attacks. We finally introduce the notion of differential immunity and show that this metric can capture the informal notion of transferability of attacks. We discuss how this measure can give us an understanding about how effective MTDeep will be. Finally, we talk about the effects of setting an incorrect $\alpha$ (that a user needs to input) when calculating the switching strategy for MTDeep.
	
	\input{figs/f_n_mnist.tex}
	
	\subsection{MTDeep as a Standalone Defense Technique}
	
	We compare the effectiveness of MTDeep with two baselines. The first one measures the accuracy of each individual network in the ensemble and the second one is a randomized ensemble that uses Uniform Random Strategy (MTD-URS) to pick one of the constituent networks with equal probability. In contrast, MTDeep uses the Stackelberg equilibrium strategy of the defender to pick a constituent DNN at random. We do not showcase comparison against deterministic (such as majority-voting or weighted) ensembles because, as discussed in the related work section (Sec. \ref{sec:rw}), these deterministic functions are equivalent to a final layer of a large network with multiple sub-componets built using CNN, RNN and MLP building blocks. None the less, to drive the point home, under the heading of differential immunity, we empirically demonstrate that majority voting ensembles obtain a lower accuracy on adversarial examples when compared to MTDeep (and even MTD-URS) for MNIST.
	
	\subsubsection{MNIST and Fashion-MNIST.}
	For each of the data sets, we trained three classification networks that, as stated before, were built using either Convolution layers (CNN), Multi-layer Perceptrons (MLP) or Hierarchical Recurrent layers (HRNN).
	The size of the train and test sets were $50000$ and $10000$ respectively.
	
	We considered three attack methods for the attacker-- the Fast Gradient Based (FGM) attack (with $\epsilon=0.3$), the DeepFool (DF) attack (with three classes being considered at each step when searching for an attack perturbation), and the Projected Gradient Descent (PGD) attack (with $\epsilon: 0.3$, $\epsilon-iter: 0.05$). We develop adversarial examples for each test image based on either the loss gradient (for FGM and PGD) or the classification boundary in the feature space (for DF) corresponding to each individual network in the ensemble. For example, the adversarial examples generated using the PGD algorithm on the loss information of the CNN is termed as $PGD_c$ in Table \ref{table:MNIST_LA}. We then find the classification accuracy of each network on these adversarial examples to compute the utility values shown in the table. Note that an adversarial example developed using information about one network may not be as effective for the other networks. We find that this is especially true for attacks like DF that exploit information about a particular network's classification boundary (eg. $DF_m$ reduces the classification accuracy of MLP to $2\%$ but is hardly effective against the other two networks. Both of these are able to classify the adversarial examples correctly more than $95\%$ of the time). On the other hand, attacks that exploit the gradient signals of a particular network are somewhat effective against the other networks, i.e. have high transferability (eg. $FGM_m$ reduces the accuracy of MLP to $3.1\%$ and the accuracy of HRNN and CNN to $\approx 25\%$ and $\approx 55\%$ respectively). We observe this trend for both the MNIST and the Fashion-MNIST data-set.
	
	In Figure \ref{fig:f_n_mnist}, we plot the accuracy of a particular classification system (the objective function value), when using MTDeep \textit{vs.} any of the single constituent networks and MTD-URS as $\alpha$ varies from $0$ to $1$.
	When $\alpha=0$ and the defender ignores the possibility of playing against an adversary, and thus, the mixed strategy for MTDeep boils down to a pure strategy for selecting the most accurate classifier. In our experiments, MTDeep choose the MLP for every input test-image for the MNIST data-set and the CNN for classifying inputs drawn form Fashion-MNIST. In contrast, MTD-URS has lower classification accuracy than MTDeep because it also uses the two less accurate classifiers equal amounts of time. Given that classification accuracies for each of the constituent networks are relatively high, the difference is hard to notice in the graph.
	
	When $\alpha=1$ and the defender cares about accuracy on only adversarial examples, strong attacks like PGD for a particular network can fool it $100\%$ of the time for MNIST data classification and at least $97\%$ for Fashion-MNIST. In contrast to using individual networks, randomized selection of networks at classification time perform much better because an adversarial perturbation developed based on information from one network fails to fool other networks that may be selected at classification time.
	MTDeep achieves a classification accuracy of $24\%$ for MNIST and $25\%$ for Fashion-MNIST while MTD-URS has a classification accuracy of $\approx 20\%$ for both the data sets. The difference in classification accuracy is mainly because MTD-URS picks more vulnerable networks with equal probability. The mixed strategies for MTDeep in the case of the two data-sets are as follows.
	
	\begin{tikzpicture}
	\begin{axis}[
	xbar stacked,
	legend style={at={(1.1,0.695)}
		, anchor=north
	},
	axis line style={draw=none},
	axis y line*=left,
	axis x line*=bottom,
	font=\small,
	symbolic y coords={MNIST,Fashion-MNIST},
	ytick=data,
	/pgf/bar width=5pt,
	xmax=1.1,
	width=0.7\linewidth,
	height=0.3\linewidth,
	enlarge y limits={abs=0.8cm}
	]
	\addplot coordinates {(0.2,MNIST) (0.001,Fashion-MNIST)};
	\addplot coordinates {(0.13,MNIST) (0.499,Fashion-MNIST)};
	\addplot coordinates {(0.67,MNIST) (0.5,Fashion-MNIST)};
	\legend{MLP, CNN, HRNN}
	\end{axis}
	\end{tikzpicture}%
	
	Note that in the case of Fashion-MNIST, MLP has very low probability of being played ($\approx 0.001\%$) and the classification system is found to be the most secure when utilizing a subset of two consequent networks (i.e. CNN and HRNN). On the other hand, for classification of MNIST data, the MLP has higher probability of being played at equilibrium than the CNN-based classifier. HRNN is given equal weight as CNN for Fashion-MNIST but clearly dominates in the case of MNIST.
	
	
	\input{figs/imagenet.tex}
	
	\subsubsection{ImageNET}
	We use six different networks which have excelled on ILSVRC-2012's validation set \cite{russakovsky2015imagenet} (Table \ref{table:A}) to construct the ensemble for MTDeep.
	Since attacks like FGM, DF and PGD on these large networks have are time intensive because they need to be calculated for every single test image, we assume the adversary uses Universal Perturbations (UP) developed for each network in \cite{moosavi2016universal}, which (1) is built on top of DF and (2) have to be generated only once. These UPs were generated by ensuring that the $L_\infty$ norm of the perturbations were less than a bound $\xi = 10$ (Table \ref{table:A}). The actions of both the players and their utilities are shown in Fig. \ref{table:A}.
	
	Researchers have shown that defense mechanisms like adversarial training are ineffective against this type of attack \cite{moosavi2016universal}. Moreover, state-of-the-art defense mechanisms (c.f. discussion in related work), are still ineffective against this attack. In such cases, MTDeep is a particularly attractive approach because it can increase the robustness of the classification system even when all other defense mechanisms are ineffective.
	
	In Figure \ref{fig:imgNT}a, we plot the expected accuracy for the MTDeep along with the objective values of each of the constituent networks when the probability of an adversary type $\alpha$ varies.
	Given there are six constituent networks in the ensemble, to avoid clutter, we don't plot MTD-URS for brevity but observe that it always has $\approx 4\%$ less accuracy than MTDeep, which is a relatively high loss in accuracy given the ImageNET data-set.
	When $\alpha=0$, MTDeep uses the most accurate network (ResNet-152) that maximizes the classification accuracy.
	As adversarial inputs become more ubiquitous and thus $\alpha$ moves towards $1$, the accuracy against the perturbed inputs drops for all the constituent networks of the ensemble.  Thus, to stay protected, MTDeep switches to a mixed policy that utilizes more networks.
	
	When the system receives only adversarial samples, i.e. $\alpha=1$,  the accuracy of MTDeep is $42\%$ compared to $20\%$ for the best of the single DNN architectures.  The optimal strategy in this case is $\vec{x} = (0, 0.171, 0.241, 0, 0.401,  0.187)$ which discards some of the configurations ($VGG$-$F$ and $VGG$-$16$ in this case).
	The $22\%$ accuracy bump for modified images comes despite (i) high misclassification rates of constituent networks against Universal Perturbations, and (ii) lack of proven defense mechanisms against such attacks. 
	
	\input{figs/at_f_n_mnist.tex}
	
	
	\subsubsection*{Remark.} Let us denote accuracy on legitimate samples as $a_L$ and accuracy on adversarial samples as $a_A$. Note that the objective function ($O$), becomes the equation of a line when $a_L$ and $a_A$ are constants because $O = (a_A-a_L)*\alpha+a_L$.
	Since the values of $a_L$ and $a_A$ are constant for each constituent network, the expected accuracy ($=O$) results in a straight line with slope $(a_A-a_L)$ and intercept $a_L$. Also, as accuracy on the legitimate samples is more than accuracy on the adversarial inputs, i.e. $a_L>a_A$, the slope is negative. 
	For the MTDeep system (and also MTD-URS), the change in the accuracy values $a_A$ and $a_L$ is small ($2-4\%$ relative to the $100\%$ scale of Y-axis) as $\alpha$ varies from $0$ to $1$. Thus, the plots although non-linear, at times \textit{appear} to be linear.
	
	%
	\subsection{MTDeep as an Add-on Defense-in-depth solution}
	\label{sec:mtd_ad}
	
	We study the use of MTDeep on top of a state-of-the-art defense mechanism called Ensemble Adversarial Training (EAT) \cite{tramer2017ensemble}. EAT is an improvement on top of the adversarial training procedure in which (1) an attack algorithm is chosen, (2) perturbed images are generated using it for a particular network and (3) the generated data is used (with their correct labels) to fine tune the weights of the trained network that needs to be made more robust. Although this helps to robustly the network to an extent, higher gains in accuracy against adversarial examples can be gained by incorporating more perturbed examples in the new test set, especially the ones that are generated by attacking other networks (i.e. use all, not only the one whose parameters will be fine tuned). As more than one network is required in this defense procedure, the authors call this as Ensemble Adversarial Training even though the end product of this procedure is a single network more robust to adversarial attacks. Note that MTDeep renders itself naturally to this robustification method and also, with high probability, uses all the robust constituent networks in the ensemble at test time.
	
	Unfortunately, using EAT can only make the networks robust against attack images generated by the particular attack algorithm it used for fine-tuning and may still be vulnerable to stronger (i.e. more computationally intensive) attacks. In Fig \ref{table:AT_MNIST_A}, we show that the utility values obtained using the three constituent networks whose parameters are fine-tuned using EAT (which, in turn uses the FGM attack to generate training samples on top of the MNIST test set). Note that although there is a boost in overall accuracy against against adversarial examples generated using FGM, the other attacks (1) DF, which is generated in a very different manner compared to FGM, and (2) PDG, which represents a stronger class of attacks, are both still able to cripple the individual constituent networks. Surprisingly, even for these attacks, the EAT procedure increase the accuracy for attacks that are {\em mis-aligned}. For example, an attack PGD$_H$ generated using the model parameters of the HRNN$_{eat}$ brings down the accuracy of the HRNN$_{eat}$ network to $9\%$ whereas, it is found to be pretty ineffective against the CNN$_{eat}$ ($\approx 81\%$) and the MLP$_{eat}$ ($\approx 72\%$).
	As to why EAT helps is reducing the transferability of these attacks could be an interesting future work. In the present context, this phenomenon helps MTDeep used in conjunction to the EAT method obtain impressive accuracy gains against attack images.
	
	We highlight the results of our experiments with the fine-tuned networks on the MNIST dataset in Fig \ref{fig:at_mnist}. When $\alpha=1$, i.e. the worst case for the defender and it only gets adversarially perturbed images as inputs, the accuracy of the constituent networks are $0-4\%$ because the EAT training is using the FGM attack is ineffective against DF and PGD attacks for a particular network. On the other hand, MTDeep achieves an accuracy of $\approx 55\%$ against adversarially perturbed images because of the reduced effectiveness in terms of transferability of the attack images. Thus, we see a gain of more than $50\%$ when classifying only adversarially perturbed images.
	
	\subsection{Blackbox Attacks on MTDeep}
	
	MTDeep designs a strategy based on a set of known attacks.  Once deployed, an attacker can train a substitute network via distillation, i.e. use MTDeep as an oracle to obtain labels for the (chosen-ciphertext like) training set for the substitute network. Given that the distilled network captures information relating to the randomization at test time, we wanted to see how effective such a distillation procedure is in generating an expected network that mimics MTDeep. More specifically, if adversarial samples generated on this distilled network \cite{papernot2017practical} successfully transfer against the MTDeep ensemble.
	
	For this purpose, we used the non-adversarially trained networks for classifying MNIST data and consider the worst-case scenario where all inputs at test time are adversarially modified, i.e. $\alpha=1$.
	Note that a distilled network needs to capture both (1) the behavior of the ensemble and (2) the built-in randomization (expected classification boundary) of the MTDeep ensemble with limited training samples ($50000$, which is equal to the size of the training set for the constituent networks) in order to be effective. We notice that MTDeep has higher immunity to blackbox attacks and is able to classify attack inputs $\approx 32\%$ of the time compared to the $\approx 24\%$ accuracy against white-box attacks, as discussed in the previous sub-section. Thus, there exists a white-box attack in the attacker's arsenal that is stronger than the black-box attack we generated, thereby not affecting the defender's optimal mixed strategy.
	
	Note that even if a blackbox attack proved to be a more effective attack against the ensemble (which it may be for some other domain or vision dataset), this attack is not modeled by the defender in the original game. The defender with knowledge of such blackbox attacks can do two actions-- (1) incorporate the blackbox attack as one of the attacker's actions which in turn, might change the mixed strategy for random selection of constituent networks and (2) train the individual networks against adversarial images generated by this attack. Both of these can, in turn, lead the attacker to come up with new black-box attacks against the improved ensemble. As to how and when, if at all, this procedure leads to a stable point is another interesting future research direction.
	
	
	\input{figs/diff_imm_stats.tex}
	
	
	\subsection{Differential Immunity}
	\label{sec:di}
	
	Clearly, when an attack $u \in U$ is able to cripple all the networks $n \in N$, using MTDeep will provide no gains in robustness. In this section, we try to quantify the gains MTDeep can provide. 
	Let $E:N\times U \rightarrow [0,100]$ denote this fooling rate function where $E(n,u)$ is the fooling rate when an attack $u$ is used against a network $n$.
	Differential immunity of an ensemble $U$ against a set of known attacks $E$ against it $\delta$ can measured with just the fooling rate values as follows,%
	\begin{eqnarray}
		\delta(U,N) &=& \min_u \frac{\max_n E(n,u) - \min_n E(n,u) + 1}{\max_{n} E(n,u) + 1} \nonumber
	\end{eqnarray}
	
	If the maximum and minimum fooling rates of $u$ on a $N$ differ by a wide margin, then the differential immunity of MTDeep is higher.  This is represented in the numerator.  The denominator ensures that an attack which has high impact (or fooling rate) reduces the differential immunity of a system compared to a low impact attack even when the numerator is the same.  The $+1$ factor in the denominator of the function prevents division by zero while the $+1$ in the numerator ensures that higher values of $\max_n E(n,u)$ reduce the $\delta$ when $\max_n E(n,u) = \min_n E(n,u)$. Note that $\delta \in [0,1]$. As per this measure, the differential immunity of the various ensembles used in our experiments are higlighted in Table \ref{table:DIandAcc}.
	
	As per our expectation, we observe a general trend that the differential immunity of an ensemble in proportional to the accuracy gains obtained by MTdeep when compared to the most secure constituent network in the ensemble. Although we notice the lowest gain in case of ImageNET, note that this $20.68\%$ is substantially better absolute gain in accuracy than the $\approx 22\%$ or the  $\approx 24\%$ gain in accuracy for the Fashion-MNIST and the MNIST datasets with non-adversarially trained DNNs because the number of classes in ImageNET is $1000$ compared to $10$ for the latter two datasets. A random class selector with zero understanding of the input (provided there is no class imbalance among the adversarial inputs) can achieve $\approx 10\%$ accuracy for MNIST and Fashion-MNIST where as it can can only obtain an accuracy of $\approx 0.001\%$ for the ImageNET data-set.
	
	Note that existing measures of robustness are mostly designed for a single DNN \cite{bastani2016measuring,weng2018evaluating} and thus, do not try to incorporate the notion of transferability, i.e. to what extent is an attack designed for one network can effect another. Thus, they cannot be used to correctly measure the robustness of an ensemble. We propose \textit{differential immunity} as one of the metrics for evaluating ensembles that use any form of randomization at test time. It can be used to capture the transferability of an adversarial attack and thus, provide a reasonable measure of robustness for ensembles.
	
	\input{figs/alpha.tex}
	
	\noindent {\em Disagreement Metrics.~} In Fig. \ref{table:DiffImm}, we highlight the number of perturbed test images (total $10000$) on which $0,1,2$ or $3$ constituent DNN's classification output(s) agree with the correct class label. We conducted these experiments using the non-adversarially trained networks for MNIST classification and for brevity purposes, we only use the FGM attack method. Note that the $FGM_{C}$ is the strongest attack that can make all the $n\in N$ misclassify at least $70\%$ of the images. As generating $\delta$ can be costly at times, which needs the fooling rates for each pair $(u,n)$, one can generate the agreement metrics on a small data set to provide upper bounds for $\delta$. This provides an idea as to how using a MTDeep ensemble can increase the robustness against adversarial samples. In this case, $\delta_{MNIST} \leq 0.51$ because for the strongest attack, every network in the ensemble will misclassify (approx.) $49\%$ of the time. Also, note that a majority based ensemble can will only be able to guarantee an accuracy of $\approx 14\%$ against the FSM$_{C}$ attack because in all the other cases, only $0$ or $1$ network is able to correctly predict the correct class. In comparison, MTDeep when facing an attacker who only uses FGM attacks obtains an accuracy of $26.8\%$ against adversarially perturbed inputs.
	
	\noindent {\em Towards Differentially Immune Networks.~} Previously, authors in \cite{szegedy2013intriguing} have shown that constructing an ensemble with high $\delta$ is difficult. The authors show that ideas like partitioning the training data into disjoint sets that are then used to train different networks ($\in N$) do not make the networks differentially immune.
	This concept of an attack's potency across networks it was not specifically targeted for is defined as transferability of an attack \cite{szegedy2013intriguing} and, although informally used, is similar our notion of differential immunity.
	Fortunately, recent works highlight promising avenues that can be used to limit the transferability of attacks \cite{adam2018stochastic}. If an ensemble of such networks can be developed, as we saw in the case of DNNs for MNIST fine-tuned with EAT, MTDeep can provide significant gains as a defense technique. In scenarios where generating ensembles with high differential immunity is still difficult, MTDeep can still boost the accuracy of classifiers (in conjunction or without other state-of-the-art defense mechanisms).
	
	
	\subsection{Participation of Individual Networks.} In Figure \ref{fig:imgNT}b, we explore the participation of individual networks in the mixed strategy equilibria for MTDeep used to classify ImageNET data. The results clearly show that while it is useful to have multiple networks providing differential immunity (as testified by the improvement of accuracy in adversarial conditions),  the leveling-off of the objective function values with more DNNs in the mix does underline that there is much room for research in actively developing DNNs that can provide greater {\em differential immunity}. Note that no more than four (out of the six) networks participate in the equilibrium. An ensemble of networks with higher differential immunity equipped with MTD can thus provide significant gains in both security and accuracy.
	
	\subsection{Robustness against Miscalibrated $\alpha$}
	So far in our discussion, we have assumed that $\alpha$ (the attacker's probability) is calibrated correctly when coming up with a randomization strategy.
	But if the value of $\alpha$ is incorrect, the computed strategy
	ends up becoming sub-optimal. In Figure \ref{fig:wa}, we plot the deviation of the chosen policy (based on the assumed $\alpha$) from the optimal as the real $\alpha$ is varied $\pm 50\%$ from the one assumed.
	The BSG-framework remains quite robust (as opposed to a uniform random strategy) i.e. the accuracy is within $0-3\%$ of the optimal accuracy. 
	The robustness to $\alpha$ further highlights the usefulness of MTDeep as a meta-defense meant to work not only against adversarial attacks but also in the context of a deployed classifier that will have to deal with adversaries as well as legitimate users.

	
	
	\section{Conclusion}
	In this paper, we introduced MTDeep -- a framework inspired by Moving Target Defense in cybersecurity -- as `security-as-a-service' to help boost the security of existing classification systems based on Deep Neural Networks (DNNs). 
	We modeled the interaction between MTDeep and the users as a Bayesian Stackelberg Game, whose equilibrium gives the optimal solution to the multi-objective problem of reducing the misclassification rates on adversarially modified images while maintaining high classification accuracy on the non-perturbed images.
	We empirically showed the effectiveness of MTDeep against various classes of attacks for the MNIST, the Fashion-MNIST and the ImageNet data-sets. 
	Lastly, we demonstrated how using MTDeep with existing defense mechanisms for DNNs result in more robust classifiers and highlighted the importance of developing ensembles with higher differential immunity.
	
	
	
	\bibliographystyle{splncs04}  
	\bibliography{mtdeep}  
	
\end{document}

%% file: data/at_mnist.tex

%% file: figs/mtdeep_concept.tex
\begin{figure*}[t]
\small
\centering
\begin{tikzpicture}[
            shorten >=2pt,
            auto,
            node distance=2cm,
            state/.style={circle,inner sep=2pt}
            ]
            \draw [draw=black] (-2.8,-3.1) rectangle (1.1,2.8);
            \node[state] (rnn) [text width=1.2cm, align=center] {{\includegraphics[width=\textwidth]{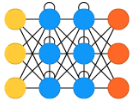}}\\{\small HRNN}};
            \node[state] (pred) [right of=rnn, text width=1cm, align=center] {{\Huge$0$}\\{\small output}};
            \node[state] (dummyRightNode1) [left of=rnn, text width=1cm, align=center] {};
            \node[state] (randomizer) [left of=rnn, text width=1.2cm, align=center] {{\Large\faRandom}\\{\footnotesize random\\selection}};
            \node[state] (dummyRightNode) [left of=randomizer, text width=0cm, align=center] {};
            \node[state] (cnn) [above of=rnn, text width=1.2cm, align=center] {{\includegraphics[width=\textwidth]{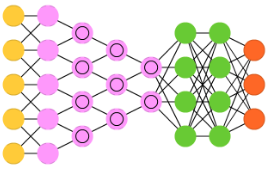}}\\{\small CNN}};
            \node[state] (mlp) [below of=rnn, text width=1.2cm, align=center] {{\includegraphics[width=\textwidth]{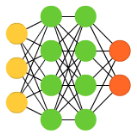}}\\{\small MLP}};
            \node[state] (img) [left of=dummyRightNode, text width=1cm, align=center]{{\includegraphics[width=\textwidth]{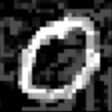}}\\{\tiny FGM$_h(i)$}};
            \node[state] (adv) [above of=img, text width=0.9cm, align=center] {{\includegraphics[width=\textwidth]{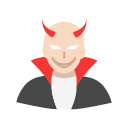}}\\{$\mathcal{A}$}};
            
            \path (adv)
                 edge  [->, Black, dashed, bend right=70]  node {$1$} (img);
            \path (img)
                 edge  [->, Black, dashed]  node {$2$} (randomizer);
            \path (randomizer)
                 edge  [->, Black, dashed, bend right=40]  node {$3$} (mlp);
            \path (mlp)
                 edge  [->, Black, dashed, bend right=40]  node {$4$} (pred);
\end{tikzpicture}
\caption{In this example, an attacker fails because (1) it chooses to perturb the image of a $0$ with an attack that works for the Hierarchical Recurrent Neural Network (HRNN) in the ensemble and (2) feeds it as input to MTDeep. MTDeep then (3) rolls a dice and (thus, randomly) picks the Multi-Layer Perceptron (MLP) to classify the input image and (4) correctly classifies the input image to a zero because the MLP was immune to the adversarial perturbation crafted by the adversary for HRNN.}
\label{fig:mtdnn}
\vspace{-10pt}
\end{figure*}

%% file: tables/gm_f_n_mnist.tex
\begin{table*}[t]
  \centering
  {\tiny
  \begin{tabular}{*{2}{c|}}
    \multicolumn{1}{c}{} & \multicolumn{1}{c}{Legitimate User ($\mathcal{L}$)}\\\hline
    \multicolumn{1}{|c|}{MTDeep} & Classification Image \\
    \hline
    \hline
 	\multicolumn{1}{|c|}{MLP} &  $99.1$\\
    \multicolumn{1}{|c|}{CNN} &  $98.3$\\
    \multicolumn{1}{|c|}{HRNN} &  $98.7$\\\cline{1-2}
  \end{tabular}
  \quad
  \begin{tabular}{|*{9}{c|}}
    \multicolumn{9}{c}{Adversarial User ($\mathcal{A}$)}\\\cline{1-9}
    $FGM_m$ & $FGM_c$ & $FGM_h$ & $DF_m$ & $DF_c$ & $DF_h$ & $PGD_m$ & $PGD_c$ & $PGD_h$\\
    \hline
    \hline
    3.1 & 20.39 & 38.93 & 1.54 & 89.8 & 93.83 & \cellcolor{Yellow!20} 0.00 & 49.00 & 61.00\\
    55.06 & 10.28 & 71.39 & 98.87 & 0.87 & 98.55 & 78.00 & \cellcolor{Yellow!20} 0.00 & 90.0\\
    25.12 & 27.24 & 11.43 & 95.38 & 83.17 & 3.66 & 23.00 & 51.00 & \cellcolor{Yellow!20} 0.00\\\cline{1-9}
  \end{tabular}
  \\[4pt]
  (a) MNIST
  \\[5pt]
  \begin{tabular}{*{2}{c|}}
    \multicolumn{1}{c}{} & \multicolumn{1}{c}{Legitimate User ($\mathcal{L}$)}\\
    \hline
    \multicolumn{1}{|c|}{MTDeep} & Classification Image \\
    \hline
    \hline
 	\multicolumn{1}{|c|}{MLP} &  $88.68$\\
    \multicolumn{1}{|c|}{CNN} &  $92.95$\\
    \multicolumn{1}{|c|}{HRNN} &  $89.16$\\\cline{1-2}
  \end{tabular}
  \quad
  \begin{tabular}{|*{9}{c|}}
    \multicolumn{9}{c}{Adversarial User ($\mathcal{A}$)}\\\cline{1-9}
    $FGM_m$ & $FGM_c$ & $FGM_h$ & $DF_m$ & $DF_c$ & $DF_h$ & $PGD_m$ & $PGD_c$ & $PGD_h$
    \\
    \hline
    \hline
    21.47 & 15.64 & 25.11 & 8.1 & 87.45 & 88.28 & \cellcolor{Yellow!20} 1.00 & 12.00 & 57.00\\
    23.42 & 6.07 & 34.76 & 88.21 & 5.37 & 92.86 & 32.00 & \cellcolor{Yellow!20} 3.00 & 61.00\\
    29.44 & 43.53 & 14.85 & 74.9 & 87.64 & 9.57 & 41.00 & 60.00 & \cellcolor{Yellow!20} 0.00\\
 	\cline{1-9}
  \end{tabular}
  \\[4pt]
  (b) Fashion MNIST
  }\vspace{5pt}
  \caption{The actions of the players and the utilities of the two user types--$\mathcal{L}$ and $\mathcal{A}$ for (a) the MNIST and (b) the Fashion-MNIST datasets. The utility of the defender is exactly the same as that of $\mathcal{L}$ for the co-operative game against the player type $\mathcal{L}$ and hundred minus the utility of $\mathcal{A}$ in the constant sum game against $\mathcal{A}$. The classification accuracy of each constituent network against the most effective attack, i.e. the worst case for each network, is highlighted in yellow.}
  \label{table:MNIST_LA}
  \vspace{-12pt}
\end{table*}%

%% file: tables/gm_imagenet.tex
\begin{table*}[t]
  \centering
  {\tiny
    \begin{tabular}{*{2}{c|}}
    \multicolumn{1}{c}{} & \multicolumn{1}{c}{$\mathcal{L}$}\\
    \hline
    \multicolumn{1}{|c|}{MTDeep} & Image \\
    \hline
    \hline
 	\multicolumn{1}{|c|}{VGG-F \cite{vggf}} & $(92.9,92.9)$\\
    \multicolumn{1}{|c|}{CaffeNet \cite{caffe}} &  $(83.6,83.6)$\\
    \multicolumn{1}{|c|}{GoogLeNet \cite{google}} &  $(93.3,93.3)$\\
    \multicolumn{1}{|c|}{VGG16 \cite{vgg16}} &  $(92.5,92.5)$\\
    \multicolumn{1}{|c|}{VGG19 \cite{vgg16}} &  $(92.5,92.5)$\\
    \multicolumn{1}{|c|}{ResNet-152 \cite{resnet}} & $(95.5,95.5)$ \\\cline{1-2}
  \end{tabular}
  \qquad
  \begin{tabular}{*{6}{|c}|}
    \multicolumn{6}{c}{Adversarial User ($\mathcal{A}$)}\\
    \hline
    $UP_{VGG-F}$ & $UP_{Caffe}$ & $UP_{GoogLe}$ & $UP_{VGG-16}$ & $UP_{VGG-19}$ & $UP_{Res}$\\
    \hline
    \hline
    \cellcolor{Yellow!20}$(6.3,93.7)$ & $(28.2,71.8)$ & $(51.6,48.4)$ & $(57.9,42.1)$ & $(57.9,42.1)$ & $(52.6,47.4)$ \\
    $(26.0,74.0)$ & \cellcolor{Yellow!20}$(6.7,93.3)$ & $(52.3,47.7)$ & $(60.1,39.9)$ & $(60.1,39.9)$ & $(52.0,48.0)$ \\
    $(53.8,46.2)$ & $(56.2,43.8)$ & \cellcolor{Yellow!20}$(21.1,78.9)$ & $(60.8,39.2)$ & $(60.2,39.8)$ & $(54.5,45.5)$ \\
    $(36.6,63.4)$ & $(44.2,55.8)$ & $(43.5,56.5)$ & \cellcolor{Yellow!20}$(21.7,78.3)$ & $(26.9,73.1)$ & $(36.6,63.4)$ \\
    $(36.0,64.0)$ & $(42.8,57.2)$ & $(46.4,53.6)$ & $(26.5,73.5)$ & \cellcolor{Yellow!20}$(22.2,77.8)$ & $(42.0,58.0)$ \\
    $(53.7,46.3)$ & $(53.7,46.3)$ & $(49.5,50.5)$ & $(53.0,47.0)$ & $(54.5,45.5)$ & \cellcolor{Yellow!20}$(16.0,84.0)$ \\\cline{1-6}
  \end{tabular}
  }\vspace{5pt}
  \caption{\small Normal form game matrices for the defender and the User types $\mathcal{A}$ and $\mathcal{L}$ for classifiers on ImageNET and corresponding  Universal Perturbation attacks. The worst case classification accuracy of each constituent networks is highlighted in yellow. Similar to Tab. \ref{table:MNIST_LA}, we notice that the attacks developed against a particular network is the most effective attack against that network.}
  \label{table:A}
\end{table*}%

%% file: figs/f_n_mnist.tex
\begin{figure*}[t!] 
\centering
    \begin{subfigure}[t]{0.48\columnwidth}
    \centering
    \resizebox{\textwidth}{!}{
        \begin{tikzpicture}[]
            \begin{axis}[
            width=1.15\linewidth,
			height=1.15\linewidth,
			xlabel={$\alpha \rightarrow$},
			xlabel style={at={(0.5, 0.05)}},
			ylabel={Accuracy $\rightarrow$},
			ylabel style={at={(0.1, 0.5)}},
			axis y line*=left,
			axis x line*=bottom,
			font=\scriptsize,
			smooth,
			mark size=1.1pt,
			mark options={solid},
			legend style ={
				at={(1,0.7)}, 
				anchor=south east,
				draw=none,
				fill=none,
				font=\tiny,
			},
			xtick=data
            ]
            \addplot [BrickRed,mark=o] table[x=prob,y=mtd_obj] {obj_mnist.dat};
            \addplot [NavyBlue,dashed,mark=star] table[x=prob,y=URS] {obj_mnist.dat};
            \addplot [Dandelion,densely dashed,mark=diamond] table[x=prob,y=CNN] {obj_mnist.dat};
            \addplot [OliveGreen,dashdotted,mark=otimes] table[x=prob,y=MLP] {obj_mnist.dat};
            \addplot [Gray,densely dotted,mark=triangle*] table[x=prob,y=HRNN] {obj_mnist.dat};
            \addlegendentry{$MTDeep$}
            \addlegendentry{$MTD$-$URS$}
            \addlegendentry{$CNN$}
            \addlegendentry{$MLP$}
            \addlegendentry{$HRNN$}
        \end{axis}
        \end{tikzpicture}
    }
    \caption{MNIST}
    \label{fig:mnist}
    \end{subfigure}%
    \hfill
    \begin{subfigure}[t]{0.48\columnwidth}
    \centering
    \resizebox{\textwidth}{!}{
        \begin{tikzpicture}[]
            \begin{axis}[
            width=1.15\linewidth,
			height=1.15\linewidth,
			xlabel={$\alpha \rightarrow$},
			xlabel style={at={(0.5, 0.05)}},
			ylabel={Accuracy $\rightarrow$},
			ylabel style={at={(0.1, 0.5)}},
			axis y line*=left,
			axis x line*=bottom,
			font=\scriptsize,
			smooth,
			mark size=1.1pt,
			mark options={solid},
			legend style ={
				at={(1,0.7)}, 
				anchor=south east,
				draw=none,
				fill=none,
				font=\tiny,
			},
			xtick=data
            ]
            \addplot [BrickRed,mark=o] table[x=prob,y=mtd_obj] {obj_fmnist.dat};
            \addplot [NavyBlue,dashed,mark=star] table[x=prob,y=URS] {obj_fmnist.dat};
            \addplot [Dandelion,densely dashed,mark=diamond] table[x=prob,y=CNN] {obj_fmnist.dat};
            \addplot [OliveGreen,dashdotted,mark=otimes] table[x=prob,y=MLP] {obj_fmnist.dat};
            \addplot [Gray,densely dotted,mark=triangle*] table[x=prob,y=HRNN] {obj_fmnist.dat};
            \addlegendentry{$MTDeep$}
            \addlegendentry{$MTD$-$URS$}
            \addlegendentry{$CNN$}
            \addlegendentry{$MLP$}
            \addlegendentry{$HRNN$}
        \end{axis}
        \end{tikzpicture}
    }
    \caption{Fashion-MNIST}
    \label{fig:fmnist}
	\end{subfigure}
    \caption{Accuracy of MTDeep with non-adversarially trained networks against (1) each of the constituent networks and (2) a uniform random strategy for randomly selecting a constituent network at test time. The gray line at the $10\%$ mark denotes the accuracy of randmonly guessing a class given an input image.}
    \label{fig:f_n_mnist}
    \vspace{-15pt}
\end{figure*}

%% file: figs/imagenet.tex
\begin{figure*}[t!] 
\centering
    \begin{subfigure}[t]{0.48\columnwidth}
    \centering
    \resizebox{\textwidth}{!}{
        \begin{tikzpicture}[]Table showing u
            \begin{axis}[
            width=1.15\linewidth,
			height=1.15\linewidth,
            xlabel={$\alpha \rightarrow$},
			xlabel style={at={(0.5, 0.05)}},
			ylabel={Accuracy $\rightarrow$},
			ylabel style={at={(0.1, 0.5)}},
			axis y line*=left,
			axis x line*=bottom,
			grid style={dashed,red},
			font=\tiny,
			smooth,
			mark size=1.1pt,
			mark options={solid},
            legend style={
                at={(0.5,0.01)}, 
				anchor=south east,
				draw=none,
				fill=none,
				font=\tiny
            },
            xtick=data
            ]
            \addplot [BrickRed,mark=o] table[x=prob,y=mtd_obj] {obj.dat};
            \addplot [NavyBlue,dashed,mark=star] table[x=prob,y=VGG-F] {obj.dat};
            \addplot [Dandelion,densely dashed,mark=diamond] table[x=prob,y=CaffeNet] {obj.dat};
            \addplot [OliveGreen,dashdotted,mark=otimes] table[x=prob,y=GoogLeNet] {obj.dat};
            \addplot [Gray,densely dotted,mark=triangle*] table[x=prob,y=VGG-16] {obj.dat};
            \addplot [Black,dotted,mark=triangle*] table[x=prob,y=VGG-19] {obj.dat};
            \addplot [RubineRed,loosely dashdotted,mark=triangle*] table[x=prob,y=ResNet-152] {obj.dat};
            \addlegendentry{$MTDeep$}
            \addlegendentry{$VGG-F$}
            \addlegendentry{$CaffeNet$}
            \addlegendentry{$GoogLeNet$}
            \addlegendentry{$VGG-16$}
            \addlegendentry{$VGG-19$}
            \addlegendentry{$ResNet-152$}
        \end{axis}
        \end{tikzpicture}
    }
    \caption{(a) Accuracy of classifiers under attack.}
    \label{fig:imagenet}
    \end{subfigure}%
    \hfill
    \begin{subfigure}[t]{0.48\columnwidth}
    \centering
    \resizebox{\textwidth}{!}{
        \begin{tikzpicture}[]
        \begin{axis}[
          	width=1.15\linewidth,
			height=1.15\linewidth,
            xlabel={Number of Neural Networks in MTDeep},
			xlabel style={at={(0.5, 0.05)}},
			ylabel={Accuracy $\rightarrow$},
			ylabel style={at={(0.1, 0.5)}},
			axis y line*=left,
			axis x line*=bottom,
			grid style={dashed,red},
			font=\tiny,
			smooth,
			mark size=1.1pt,
			mark options={solid},
            legend style={
                at={(0.95,0)}, 
				anchor=south east,
				draw=none,
				fill=none,
				font=\tiny
            },
            xtick=data
        ]
        \addplot [BrickRed,mark=o] table[x=prob,y=ov0] {varyN.dat};
        \addplot [NavyBlue,dashed,mark=star] table[x=prob,y=ov25] {varyN.dat};
        \addplot [Dandelion,densely dashed,mark=diamond] table[x=prob,y=ov50] {varyN.dat};
        \addplot [OliveGreen,dashdotted,mark=otimes] table[x=prob,y=ov75] {varyN.dat};
        \addplot [Gray,densely dotted,mark=triangle*] table[x=prob,y=ov1] {varyN.dat};
        \addlegendentry{$\alpha=0$}
        \addlegendentry{$\alpha=0.25$}
        \addlegendentry{$\alpha=0.5$}
        \addlegendentry{$\alpha=0.75$}
        \addlegendentry{$\alpha=1$}
    \end{axis}
    \end{tikzpicture}
    }
    \caption{(b) Participation of constituent networks.}
    \label{fig:22}
	\end{subfigure}
    \caption{For the ImageNET dataset, we (a) compare the accuracy of MTDeep {\em vs.} the constituent networks and (b) analyze the participation of the different constituent networks at equilibrium for different values of $\alpha$.}
    \label{fig:imgNT}
    \vspace{-14pt}
\end{figure*}

	
        

%% file: figs/at_f_n_mnist.tex
\begin{figure*}[t!]
\tiny
\centering
\begin{minipage}{.59\textwidth}
\centering
  \begin{tabular}{*{2}{c|}}
    \multicolumn{1}{c}{} & \multicolumn{1}{c}{Legitimate User ($\mathcal{L}$)}\\\hline
    \multicolumn{1}{|c|}{MTDeep} & Classification Image \\
    \hline
    \hline
 	\multicolumn{1}{|c|}{MLP$_{eat}$} &  $97.99$\\
    \multicolumn{1}{|c|}{CNN$_{eat}$} &  $98.97$\\
    \multicolumn{1}{|c|}{HRNN$_{eat}$} &  $97.22$\\\cline{1-2}
  \end{tabular}
  \\[5pt]
  \begin{tabular}{|*{9}{c|}}
    \multicolumn{9}{c}{Adversarial User ($\mathcal{A}$)}\\\cline{1-9}
    $FGM_m$ & $FGM_c$ & $FGM_h$ & $DF_m$ & $DF_c$ & $DF_h$ & $PGD_m$ & $PGD_c$ & $PGD_h$\\
    \hline
    \hline
    95.06 & 75.32 & 70.1 & 1.5 & 96.97 & 95.73 & 0.00 & 88.00 & 69.00 \\
    61.44 & 96.55 & 68.58 & 98.36 & 0.79 & 96.09 & 72.00 & 20.00 & 81.00 \\
    81.24 & 84.79 & 93.1 & 96.85 & 95.9 & 4.41 & 82.00 & 71.00 & 10.00 \\
    \hline
  \end{tabular}
  \caption{\small The utilities for the players when the adversary uses the aforementioned attacks against the classifiers fine-tuned using Ensemble Adversarial Training (EAT) with FGM attacks.}
  \label{table:AT_MNIST_A}
\end{minipage}
\hfill
\begin{minipage}{.37\textwidth}
\resizebox{\textwidth}{!}{
	\begin{tikzpicture}[]
        \begin{axis}[
            width=1.05\linewidth,
			height=1.05\linewidth,
            xlabel={$\alpha \rightarrow$},
			xlabel style={at={(0.5, 0.05)}},
			ylabel={Accuracy $\rightarrow$},
			ylabel style={at={(0.1, 0.5)}},
			axis y line*=left,
			axis x line*=bottom,
			grid style={dashed,red},
			font=\tiny,
			smooth,
			mark size=1.1pt,
			mark options={solid},
            legend style={
                at={(0.6,0.01)}, 
				anchor=south east,
				draw=none,
				fill=none,
				font=\tiny
            },
            xtick=data
        ]
        \addplot [BrickRed,mark=o] table[x=prob,y=mtd_obj] {obj_at_mnist.dat};
        \addplot [Dandelion,densely dashed,mark=diamond] table[x=prob,y=CNN] {obj_at_mnist.dat};
        \addplot [OliveGreen,dashdotted,mark=otimes] table[x=prob,y=MLP] {obj_at_mnist.dat};
        \addplot [Gray,densely dotted,mark=triangle*] table[x=prob,y=HRNN] {obj_at_mnist.dat};
        \addlegendentry{$MTDeep$}
        \addlegendentry{$CNN$}
        \addlegendentry{$MLP$}
        \addlegendentry{$HRNN$}
    \end{axis}
    \end{tikzpicture}
}
\vspace{-11pt}
\caption{\small Accuracy of MTDeep with adversarially trained networks.}
\label{fig:at_mnist}
\end{minipage}
\vspace{-12pt}
\end{figure*}%

%% file: figs/diff_imm_stats.tex
\begin{table}[t]
    \centering
    \begin{tabular}{|l|c|c|c|c|}
    \hline
    Networks     & \begin{tabular}[c]{@{}c@{}}Differential\\ Immunity ($\delta$)\end{tabular} & \begin{tabular}[c]{@{}c@{}}Accuracy of Best\\ Constituent Net\end{tabular} & \begin{tabular}[c]{@{}c@{}}Accuracy of\\ MTDeep\end{tabular} & Gain      \\
    \hline
    \hline
    FashionMNIST & $0.11$                                                                     & $3\%$                                                                      & $24.8\%$                                                     & $21.8\%$  \\
    MNIST        & $0.19$                                                                     & $0\%$                                                                      & $23.68\%$                                                    & $23.68\%$ \\
    ImageNET     & $0.34$                                                                     & $22.2\%$                                                                   & $42.88\%$                                                    & $20.68\%$ \\
    MNIST + EAT  & 0.78                                                                       & $4.41\%$                                                                   & $54.71\%$                                                    & $50.3\%$  \\ \hline
    \end{tabular}
    \vspace{5pt}
    \caption{Differential Immunity of the various ensembles and the gains seen in accuracy compared to the best constituent networks when $\alpha=1$.}
    \label{table:DIandAcc}
    \vspace{-20pt}
\end{table}


%% file: figs/alpha.tex
\begin{figure*}[t!]
\small
\centering
\begin{minipage}{.385\textwidth}
    \begin{tabular}{|*{5}{c|}}
        \hline
        Attacks & 0 & 1 & 2 & 3 \\
        \hline
        $FGM_C$ & 4788 & 3641 & 1449 & 118 \\
        $FGM_H$ & 389 & 2728 & 6667 & 212 \\
        $FGM_M$ & 1513 & 5790 & 2479 & 214 \\
        $FGM_{BB}$ & 2305 & 2569 & 2678 & 2444 \\
        \hline
    \end{tabular}
    \caption{Agreement among constituent networks when classifying perturbed inputs for the MNIST data-set.}
    \label{table:DiffImm}
\end{minipage}
~
\begin{minipage}{.57\textwidth}
    \centering
    \begin{tikzpicture}[]
        \begin{axis}[
            width=0.93\linewidth,
			height=0.5\linewidth,
			xlabel={$\%$ deviation of $\alpha \rightarrow$},
			ylabel={Opt $-$ Accuracy $\rightarrow$},
			axis y line*=left,
			axis x line*=bottom,
			font=\tiny,
			smooth,
			mark size=1.1pt,
			mark options={solid},
			legend style = {
				at={(0.6,0.25)}, 
				anchor=south west,
				draw=none,
				fill=none,
				font=\tiny,
			},
			xtick=data
            ]
            \addplot [BrickRed,mark=o] table[x=alpha,y=mtdeep] {attacker_robustness.dat};
            \addplot [NavyBlue,dashed,mark=star] table[x=alpha,y=mtd_urs] {attacker_robustness.dat};
            \addlegendentry{$MTDeep$}
            \addlegendentry{$MTD$-$URS$}
        \end{axis}
        \end{tikzpicture}
    
    \caption{Loss in accuracy when real world $\alpha$ is different from the $\alpha$ MTDeep uses for modeling.}
    \label{fig:wa}
    \vspace{-15pt}
\end{minipage}
\end{figure*}